\documentclass[letterpaper, 10 pt, conference]{ieeeconf}  

\IEEEoverridecommandlockouts                              

\usepackage{graphics} 
\usepackage{epsfig} 
\usepackage{mathptmx} 
\usepackage{times} 
\usepackage{amsmath} 
\usepackage{amssymb}  
\usepackage{comment}
\usepackage{amsfonts,bbm} 
\usepackage{soul}
\usepackage{xcolor}
\usepackage{subcaption}
\usepackage{multirow}
\usepackage[normalem]{ulem}
\useunder{\uline}{\ul}{}
\usepackage[table,xcdraw]{xcolor}
\usepackage{censor}
\usepackage{balance}
\usepackage{hyperref}

\newcommand{\review}[1]{{\color[rgb]{0.0,0.0,0.0} #1}}
\newcommand{\last}[1]{{\color[rgb]{0.0,0.0,0.0} #1}}
\StopCensoring
\title{\LARGE \bf
Swarm sign language: motion-based communication between drones}

\author{\censor{Thomas Rey$^{1}$, Julien Moras$^{1}$, Alexandre Eudes$^{1}$ and Antoine Manzanera$^{2}$}
\thanks{This work has been partially supported and carried out in the framework of the \censor{VORTEX project (ANR-23-IAS2-0005-01)}.}
\thanks{$^{1}$\censor{are with DTIS, ONERA, Université Paris-Saclay, 6 Chemin de la Vauve} \censor{aux Granges, Palaiseau, 91120, France}
        {\tt\small \censor{name.surname@onera.fr}}}%
\thanks{$^{2}$\censor{is with the U2IS, ENSTA, Institut Polytechnique de Paris}, \censor{828 Boulevard des Mar\unexpanded{é}chaux,  Palaiseau, 91120, France}
        {\tt\small \censor{antoine.manzanera@ensta.fr}}}%
}

\begin{document}

\maketitle
\thispagestyle{empty}
\pagestyle{empty}

\begin{abstract}
\review{In stealth-constrained swarm robotics, visual communication provides a critical alternative to active radio transmissions, which might be jammed. This research investigates motion-based communication for non-active information exchange, utilizing modular, dynamically feasible planar trajectories as visual cues. On the receiver drone end, a pose estimator tracks the transmitting drone's pose, feeding it into our custom 3DTrajDecoder. The decoder is designed to classify and segment the spatiotemporal sequence while simultaneously regressing its size and normal vector. To robustly train the decoder on both communicative and non-communicative trajectories, we developed a confi\-gurable online procedural generation pipeline. We validate our system through real-world testing and simulation to define its operating domain, supported by an extensive ablation study detailing our architectural choices and system limitations.}

\end{abstract}

\section{INTRODUCTION}
In recent years, the burgeoning field of swarm robotics has undergone major advancements, with drone swarms being deployed for diverse applications such as 
search and rescue and worker safety monitoring \cite{SaskaGestureHSI2025}. Efficient communication among drones is paramount to achieve coordinated behaviors and successful mission execution. While traditional communication methods (e.g., radio frequency (RF)) are the standard, they are susceptible to interference, jamming, or range limitations in contested environments \cite{yu2025electronic}.
To address these vulnerabilities, this paper explores the largely unexplored territory of motion-based communication (MBC) using RGB cameras - sensors ubiquitous on unmanned aerial vehicles (UAVs) - as the primary medium for 
communication. By leveraging the visual spectrum, we aim to develop a novel, MBC protocol that enhances the autonomy, stealth, and resilience of drone swarms operating in dynamic environments.

\begin{figure}[t!]
     \centering
     \includegraphics[width=\linewidth]{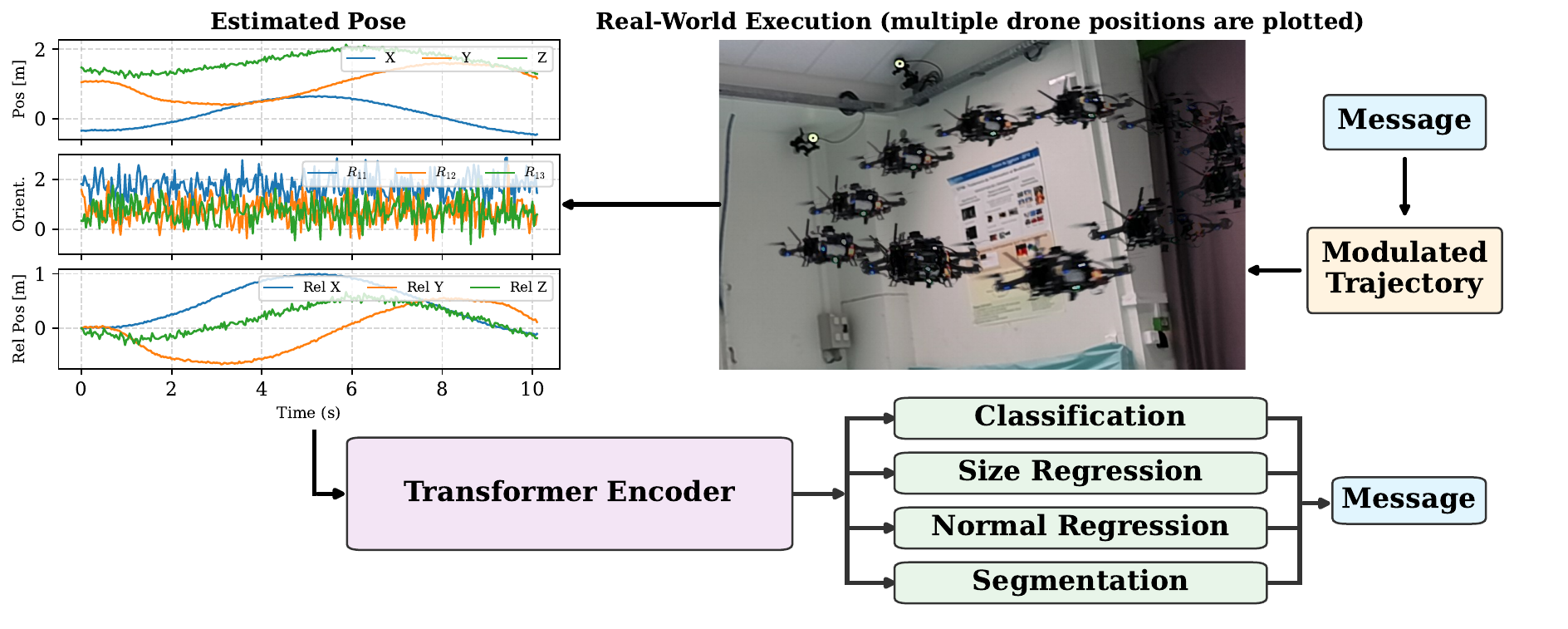}
     \caption{\review{A transmitter drone wants to send a message, through a modulated trajectory. A receiver drone estimates the transmitter's pose, and, using our 3DTrajDecoder, decodes the sent message.}}
     \label{fig: teaser}
 \end{figure}

\review{While the application of MBC to UAV swarms is quite underexplored, the underlying principles draw inspiration from a broad spectrum of established disciplines. Visual communication itself is a multifaceted domain, ranging from natural phenomena like the bee dance to engineered solutions such as UV-based optical channels \cite{9836151}. To design a robust system for drones, we look specifically to fields where movement conveys meaning. The Sign Language Recognition and Production field offers the closest parallel to our objectives. Just as recent advancements use Transformers and diffusion models to translate human gestures into semantics \cite{baltatzis2024neuralsignactorsdiffusion}. Research in Human-Robot Interaction highlights the importance of "kinemes" and interpretable gestures \cite{10.1145/3495245}.} \last{While Action Recognition provides the computer vision foundation - using 3D CNNs and spatiotemporal reasoning - to classify motion patterns from video sequences \cite{10143200}. Since drones operate in three-dimensional space, we also draw on insights from GeoAI, which tackles the challenge of encoding trajectory and intention within a geospatial context \cite{wu2025torchspatiallocationencodingframework}.}
\review{By synthesizing these interdisciplinary insight,} our study introduces a framework\review{, shown on the figure \ref{fig: teaser},} where communication occurs between a transmitting and a receiving UAV strictly through observed flight trajectories. The transmitter encodes messages by executing specific movements selected from a designed alphabet of distinct planar geometric forms (e.g., squares, circles, spirals). To maximize the information density of a single gesture, the message is modulated by the trajectory's physical scale and the normal vector of its execution plane. The receiver employs a pose estimator to track the transmitter, feeding the extracted spatiotemporal sequence into our proposed 3DTrajDecoder, a Transformer-based architecture designed to classify the shape, segment the spatiotemporal sequence and regress its continuous geometric properties.

\review{Our primary contributions are: a modulable motion-based communication for UAV-swarms - using planar primitives modulated by scale and orientation - to transmit complex semantics without RF signals.
A multi-task Vision Transformer inspired architecture capable of simultaneously classify, segment and regress both size and normal of a noisy spatiotemporal sequence. A configurable online generation pipeline, for synthesising dynamically feasible continuous trajectories, bridging the reality gap via structured controller and observation noise injection. Also, an extensive evaluation on synthetic data highlighting several design choices and limits of this system. We demonstrate the physical viability of the proposed protocol through real-world experiments. }
 \begin{figure*}
 \vspace{0.2cm}
     \centering
     \includegraphics[width=1.0\linewidth]{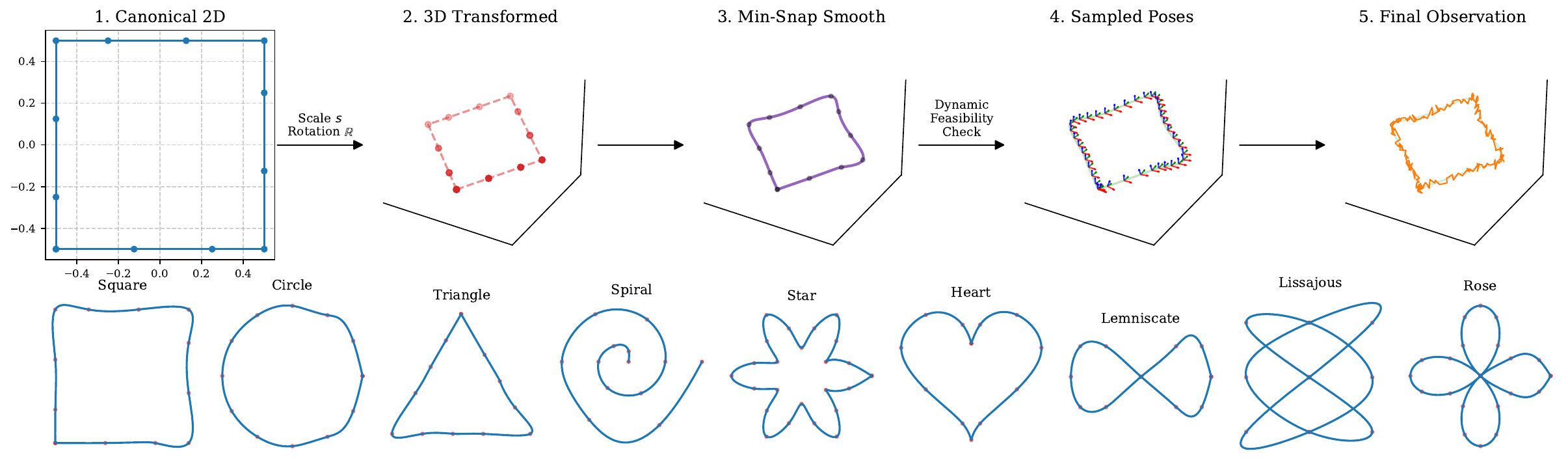}
     \caption{Dataset pipeline generation and all the possible symbols with their minimum-snap trajectories}
     \label{fig: dataset pipeline generation}
 \end{figure*}

\section{Related works}

Gestures are a means of communication, in the animal kingdom or within human interaction. It is shown in multiple studies that bees, ants, birds or fish use gestures to provide information, whether it is to control the flock or give the position of the pollen \cite{Animal2Swarm2023}. It turns out that gestures are a robust alternative to electromagnetic communication that can be either unusable (e.g. underwater), or jammed. While early methods focused on the control-theory foundations of signaling through motion, recent advancements have been done using deep learning frameworks and formal grammatical structures.

To our knowledge, the first article \cite{5354808} that directly uses relative motion between robots to communicate, focuses on the control strategies to encode information into the trajectory of an nonholonomic vehicles, by modulating their relative distance and orientation. \last{\cite{baillieul2011controltheorymotionbasedcommunication} decompose the Salsa dance into four motion primitives (steps) and model communication as a dual-objective optimal control problem - moving while overlaying on top a communication signal - they make a first formal distinction between communicative and non-communicative movements.} While most earlier works assumed a perfect observation, some further research addressed the challenge of perception. The authors of \cite{nishimura2018active}, use the alphabet letter as communicative sequences. Using a monocular camera and an Extended Kalman-Filter the authors classify the seen letter and estimate the relative pose between the two communicative drones. Moreover, the receiver drone autonomously moves to viewpoints minimizing the entropy of the seen-trajectory class distribution, as for example the letter L and V can appear identical from different viewpoints on the receiver image plane. The communicative movements used in this study were quite large: over 2m of radius. As earlier works show, the use of an alphabet composed of kinemes - i.e., movements primitives functioning as a gestural language - is mandatory so as to communicate between two robots. Fulton et al. in \cite{10.1145/3495245}, did a grounding work of creating a general kinemes' alphabet: the Robot Communication Via Motion, and implemented it on multiple platforms such as wheeled-robots, AUVs and UAVs. Their alphabet is created so to achieve Human-Robot interaction. These movements are then constrained to be easily understandable by any human and resilient to viewing angle. \cite{enan2022roboticdetectionhumancomprehensiblegestural} extended this work to communicate directly between multiple AUVs (so that humans can still understand the movements). \last{Unlike the probabilistic filtering used by \cite{nishimura2018active},} this technique uses an attention-based neural network, that directly takes images from an RGB camera to extract spatiotemporal features to classify the image sequences. 

\last{Moving beyond holistic trajectory - where one movement equals one complete message - recent Human-Robot Interaction study \cite{vanc2023communicating}, proposed a communication system that combines multiple gestures types - e.g. action, deictic and metric parameters - into a full gesture sentence. Allowing humans to express complex intents, from sequence of hand movements to robot action, using behavior trees. Still some non-neural networks based methods are explored such as Albergo et al. in \cite{albergo2025motion} which use a model-free approach, dense optical flow and directional shifts in motion to encode the messages using a robotic arm.}

\section{Contributions}

As we aim to produce a secondary MBC channel, to be use when traditional ones are unusable and consistent with prior work on aerial gestures \cite{SaskaGestureHSI2025,10.1145/3495245}, we define 9 different kinemes based on fundamental 2D geometric primitives, which can all be seen on the figure \ref{fig: dataset pipeline generation}. To enhance expressivity, messages are encoded not only by the shape class, but also by geometric modifiers: we utilize the scale and the plan orientation to encode semantic information such as a float or a directionality. For example, in "let's take the second right corridor", "let's take the corridor" can be expressed with the square, "second" with its size and "right" with the normal pointing towards the right.
To overcome the data scarcity inherent in deep-learning for robotics, we developed a procedural generation pipeline capable of synthesizing diverse dynamically feasible 3D trajectories of communicative and non-communicative movements. \review{This pipeline creates samples $X \in \mathbb{R}^{L_{max} \times D}$, where $L_{max}$ denotes the number of discrete samples in the trajectory and $D$ the dimension of the drone's state space.} The generation process is divided into four stages, as seen on the figure \ref{fig: dataset pipeline generation}: geometric waypoints definition, scale and orientation, temporal optimization and realistic noise injection.

\subsection{Architecture}

To decode symbols directly in trajectory space, we propose a multi-task Transformer architecture adapted for 3D pose sequences. Inspired by the Vision Transformer (ViT), our model treats temporal trajectory segments as analogous to image patches. The processing pipeline consists of three stages. \review{ First, to mitigate the quadratic complexity of self-attention $(O(L_{max}^2))$, the input pose sequence $X\in\mathbb{R}^{L_{max} \times D_i}$ - $D_i$ can be either or a combination of the position, the orientation (parametrize as R6D~\cite{zhou2019continuity}) or the relative position (position w.r.t. the first one) - is first compressed by a factor of $k=4$ via a 1D convolutional layer.} This downsampled representation is augmented with a learnable [CLS] token and sinusoidal positional (i.e., temporal) encodings \cite{vaswani2017attention}. The [CLS] token serves as a global query, aggregating information across the entire sequence through the self-attention mechanism to form a compact latent representation of the trajectory. Then the augmented sequence is processed by a stack of N Transformer encoder layers. These layers utilize multi-head self-attention to capture long-range temporal dependencies and refine the latent features. At the end the model uses four different task-specific heads based on the nature of the prediction task, either global over the full trajectory: shape classification, size and normal regression, or dense: message segmentation. The global tasks solely use the final state of the [CLS] token. This global vector is fed into separate Multi-Layer Perceptrons (MLPs) one per task. For temporal segmentation, the model utilizes the sequence of feature tokens. To recover the original temporal resolution, these features are upsampled via a transposed convolution layer before final classification.

\subsection{Creation of the synthetic dataset}

To train the 3DTrajDecoder, there is a need to generate thousands of dynamically feasible trajectories mimicking our predefined alphabet and some non-communicative trajectories. The dataset generation is shown on the figure \ref{fig: dataset pipeline generation}. For a given kineme, a sequence of base waypoints $\mathcal{W}_{base}$ is generated in a canonical 2D frame (step 1). These are scaled by $s$, rotated into the target 3D plane defined by $\mathcal{R}\in SO(3)$, and translated to a spatial center (step 2). To ensure dynamic feasibility, we treat these waypoints as constraints in a Minimum Snap optimization problem. We solve it for a piecewise polynomial trajectory of degree 8, minimizing the fourth derivative (snap) to produce smooth, continuous motions, as described in \cite{Mueller2015}\footnote{\review{The minimum-snap code can be retrieved at: \url{https://pypi.org/project/minsnap-trajectories/}}}. The time allocation for each segment is proportional to the Euclidean distance between waypoints, scaled by a total execution time $T_{exec}$ sampled from a uniform distribution $T_{exec} \sim \mathcal{U}(T_{min}, T_{max})$. We bridge the reality gap by injecting a controller noise: before trajectory optimization, Gaussian noise is added to the waypoints. This simulates the drone's physical inability to track a perfect geometric path.

Real-world gesture sequences are never isolated. To simulate realistic behavior, we embed the communicative trajectory within a continuous stream of motion. We generate "non-communicative" segments using a bounded 3D random walk algorithm. To prevent kinematic discontinuities, strictly static hovering phases (duration 0.5s) are inserted between the random walk and the communicative gesture at the start and the end. Boundary constraints (zero velocity and acceleration) are enforced at these junctions to ensure smooth transitions. \review{As the expected 3DTrajDecoder input length is $L_{max}$ - fixed at $L_{max}$ = $T_{max}$ x $F$, $F$ being the sampling rate of the estimated input sequence - the continuous stream of motion must be at least $T_{max}$ seconds.}

Finally, every generated sample undergoes a dynamic feasibility check. Trajectories requiring thrust or body rates exceeding the physical limits of a standard quadrotor platform are discarded and regenerated (step 3). \review{While the  class distribution is uniform, differences in dynamic feasibility among classes cause unequal rejection rates. Ultimately resulting in a non-uniform size distribution.} 

As the receiver uses the sequence of pose estimated by a pose estimator as an input for the model, an observation noise is added: the continuous trajectory is sampled at $F$~Hz into a sequence of length $L_{max}$ (step 4); we then apply an observation noise to the state estimates, mimicking the degradation of the pose estimation system. 

To address the need for diverse and adaptable training and testing scenarios, our procedural generation pipeline is flexible and online. Unlike static datasets, our system synthesizes 3D trajectories on-the-fly based on a modular configuration file. This approach allows for rapid experimentation with various environmental parameters, such as noise intensity, trajectory scale, and class distribution, without the overhead of storing massive datasets. To ensure scientific rigor and reproducibility, all stochastic components (including random walks and noise injection) are initialized with fixed seeds, ensuring that any specific dataset configuration can be exactly recreated.

\subsection{Creation of an image dataset}

\begin{figure}[ht!]
     \centering
     \begin{subfigure}[b]{0.49\linewidth}
         \centering
         \includegraphics[width=0.95\linewidth]{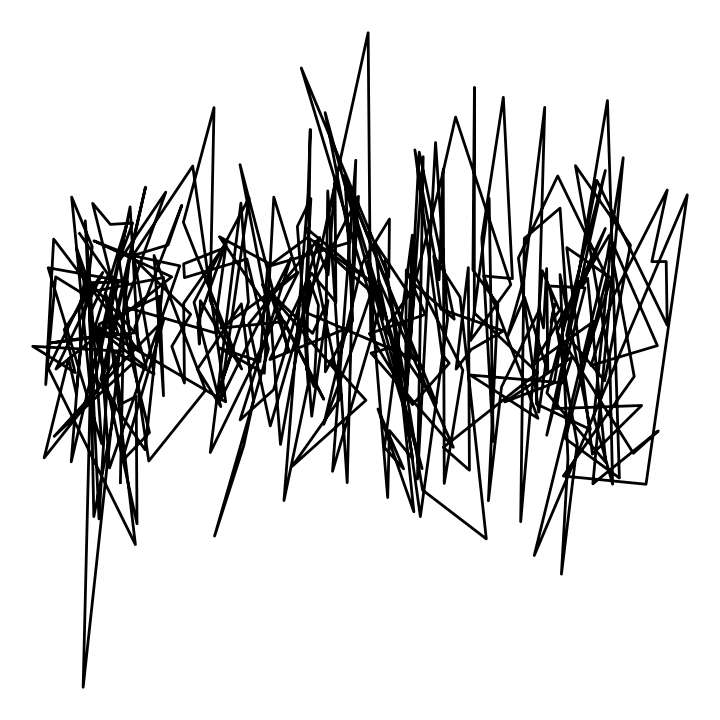}
         \caption{Class: Lemniscate}
     \end{subfigure}
     \hfill
     \begin{subfigure}[b]{0.49\linewidth}
         \centering
         \includegraphics[width=0.95\linewidth]{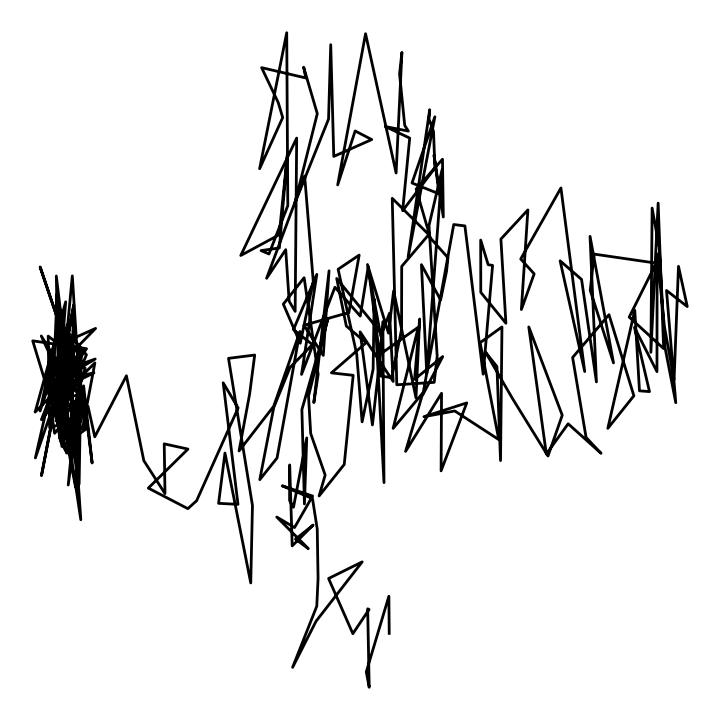}
         \caption{Class: Rose}
     \end{subfigure}
        \caption{
        Two sample images of projected trajectories to be tested on image classification models (resolution $740 \times 740$).
        }
        \label{fig: Image classification}
\end{figure}

As no real state of the art exists, since no one uses the same alphabet, no dataset is publicly accessible. It is then hard to compare solely on accuracy. Moreover, our approach tries to predict way more information than solely a movement class. So for comparison purposes, we updated our dataset creation pipeline to save trajectories as images. The image is created using either the ground-truth normal or the estimated one, which is used as the normal of the image plane viewed from the observer. The distance is not taken into consideration so that all trajectories fit the whole image support. Some example images can be seen on the figure \ref{fig: Image classification}. Two classic architectures - a ResNet32 \cite{he2016deep} and a ViT \cite{vaswani2017attention} - are then trained on a classification computer vision task - using  the same classification loss used in the 3DTrajDecoder - which only aims to predict the class of the seen movements in the image.

\subsection{Evaluation Metrics}
To evaluate the performance of our models, we employ various metrics: 

\subsubsection{Geodesic Error ($\mathcal{E}_G$)} 
Measures the distance between the predicted trajectory normal $\hat{N}$ and the ground truth normal $N$ on the manifold. For points on a unit sphere, this is defined as:
\begin{equation}
    \mathcal{E}_G(N, \hat{N}) = \arccos(N \cdot \hat{N})
\end{equation}
\subsubsection{Signal-to-Noise Ratio (SNR)} 
Quantifies the signal strength relative to the trajectory radius, i.e., the maximum distance between two points of the kineme.
The SNR is defined as the ratio of the perfect kineme's radius over the radius of the noisy one (i.e., the observed one).
\subsubsection{Completeness ($\Phi_r$)} 
Represents the proportion of the kineme, which is in the input sequence. Letting $T_{exec}$ be the duration of the kineme, $N_m$ the kineme label, $N_i$ the $i^{th}$ sample label, $F$ the sampling frequency of the input sequence and $\mathbbm{1}(\cdot)$ the indicator function.
\begin{equation}
    \Phi_r = \frac{1}{T_{exec}F} \sum_{i=1}^{L} \mathbbm{1}(N_i = N_m)
\end{equation}

\review{Different metrics are also used such as the F1-score for both classification: $F_1^{\text{\tiny classif}}$ and segmentation: $F_1^{\text{\tiny seg}}$. The mean absolute and square errors (MAE, MSE) are used to indicate the performance of the size regression head. For the normal regression head, the mean $\mathcal{E}_G$, the median $\mathcal{E}_G$ and the accuracy for two thresholds (10° and 30°) are used.}

\subsection{Training parameters}
Prior to being used as input, the position sequence is normalized by using the extrema of the sampling distribution for each axis to center the data around the origin. The model is optimized using a multi-objective loss function: Cross-Entropy for classification and segmentation, L1 loss for size regression. We employ the Adam optimizer with a ReduceLROnPlateau scheduler. For each training, an early-stop is used and calculated on the following weighted sum of metrics for each head:
\begin{equation}
    \frac{1}{4} \big(F_1^{\text{\tiny classif}}+F_1^{\text {\tiny seg}}+\max(1-\frac{MAE}{5},0) +\max(1-\frac{\mathcal{E}_G}{20^\circ},0) \big)
\end{equation}
To enforce geometric consistency in the prediction of the 3D normal, we used a dot-product based loss ($\mathcal{L}_{dot}$):

\begin{equation}
\mathcal{L}_{dot} = 1 - \hat{N} \cdot N \label{eq:loss_dot}
\end{equation}

\section{Results}
\last{
As our dataset pipeline generation is completely modular, we define for this section a specific system tailored to the embedded constraints. To enable dynamic communication between drones, we fixed a 12s input window for the Transformer to be able to have both fast kineme ($T_{min}$ = 3s) and slow ones ($T_{max}$ = 12s). The goal is to have a single model embedded in the receiver drone and still be able to communicate fast when the transmitter drone is close and slower when it is further away. These limits are also chosen so as to have dynamically feasible trajectories in small or large kinemes. 
Part of the main generation parameters can be seen in Table~\ref{tab: config datasets}, the test set follows the same configuration but with $10 \times$ fewer samples.

Since the model requires pose sequences as input, we chose a specific RGB pose estimator \cite{REY2025103339}. The field of monocular 6D pose estimation - predicting an object's 3D position and orientation from a single image - has advanced significantly through recent deep learning methods. For embedded robotics applications, such as onboard drone processing of the pose estimator, regression based method are particularly relevant due to their low inference latency \cite{stapf2023pvit}. In the following section we consider the observation noise of the RGB pose estimator \cite{REY2025103339} and its onboard sampling rate of 30Hz.}

\begin{table}[ht!]
\centering
\caption{Train set configuration to train all the models}
\begin{tabular}{|c|c|c|ccc|c|}
\hline
\multirow{2}{*}{\begin{tabular}[c]{@{}c@{}}Number\\ samples\end{tabular}} & \multirow{2}{*}{\begin{tabular}[c]{@{}c@{}}Size\\ (m)\end{tabular}} & \multirow{2}{*}{\begin{tabular}[c]{@{}c@{}}$T_{exec}$\\ (s)\end{tabular}} & \multicolumn{3}{c|}{\begin{tabular}[c]{@{}c@{}}Position\\ center (m)\end{tabular}}  & \multirow{2}{*}{$SNR_{min}$} \\ \cline{4-6}
                                                                                &                       &                              & \multicolumn{1}{c|}{x}          & \multicolumn{1}{c|}{y}          & z           &                           \\ \hline
100,000                                                                         & {[}0.01,1{]}           & {[}3,12{]}                   & \multicolumn{1}{c|}{{[}-5,5{]}} & \multicolumn{1}{c|}{{[}-5,5{]}} & {[}0.5,6{]} & 0.1                       \\ \hline
\end{tabular}
\label{tab: config datasets}
\end{table}

\subsection{Multiple models comparison}

\review{Multiple models have been evaluated:
a transformer-based architecture, an LSTM-based one, and more traditional classification architectures such as a ViT and a ResNet on the test set.}
Their performance on the test dataset is shown on Table~\ref{tab: Transformer vs LSTM vs ResNet vs ViT on the test}. \review{The performance of the Transformer and LSTM base architectures are very close, except on the segmentation task: as the LSTM has a sequential structure, its scope is limited by its local and causal memory, whereas the Transformer does have a global scope with the attention mechanism, which makes the global task - over the full sequence - easier.} Our multi-task architecture outperforms the two classic image-based models on classification, as shown on Table~\ref{tab: Transformer vs LSTM vs ResNet vs ViT on the test}, while also being trained on 3 other tasks.

\begin{table}[ht!]
\vspace{0.15cm}
\centering
\caption{Performance of the Transformer, LSTM-based approaches and the baselines. The best result is in bold and the second best is underlined.}
\begin{tabular}{|l|c|c|c|c|}
\hline
\multicolumn{1}{|c|}{{\ul \textit{On Test}}}      & Transformer    & LSTM           & ResNet & ViT   \\ \hline
$F_1^{\text{\tiny classif}}$ $\uparrow$           & {\ul 91.13}    & \textbf{92.93} & 73.79  & 73.94 \\ \hline
$F_1^{\text{\tiny seg}}$ $\uparrow$               & \textbf{94.16} & {\ul 78.44}    & X      & X     \\ \hline
MAE (m) $\downarrow$                              & \textbf{0.057} & {\ul 0.0572}   & X      & X     \\ \hline
MSE (m) $\downarrow$                              & {\ul 0.0061}   & \textbf{0.006} & X      & X     \\ \hline
Mean $\mathcal{E}_G$ (°) $\downarrow$             & \textbf{13.39} & {\ul 13.44}    & 25.23  & 25.23 \\ \hline
Median $\mathcal{E}_G$ (°) $\downarrow$           & \textbf{5.99}  & {\ul 6.34}     & 11.75  & 11.75 \\ \hline
Accuracy (\textless{}10°) $\uparrow$ & {\ul 69.7}     & \textbf{69.8}  & 39.2   & 39.2  \\ \hline
Accuracy (\textless{}30°) $\uparrow$ & {\ul 90.9}     & \textbf{91.2}  & 66.6   & 66.6  \\ \hline
\end{tabular}
\label{tab: Transformer vs LSTM vs ResNet vs ViT on the test}
\vspace{-0.25cm}
\end{table}

\begin{figure}[ht!]
\vspace{-0.3cm}
     \centering
     \begin{subfigure}[b]{0.49\linewidth}
         \centering
             \includegraphics[width=0.99\linewidth]{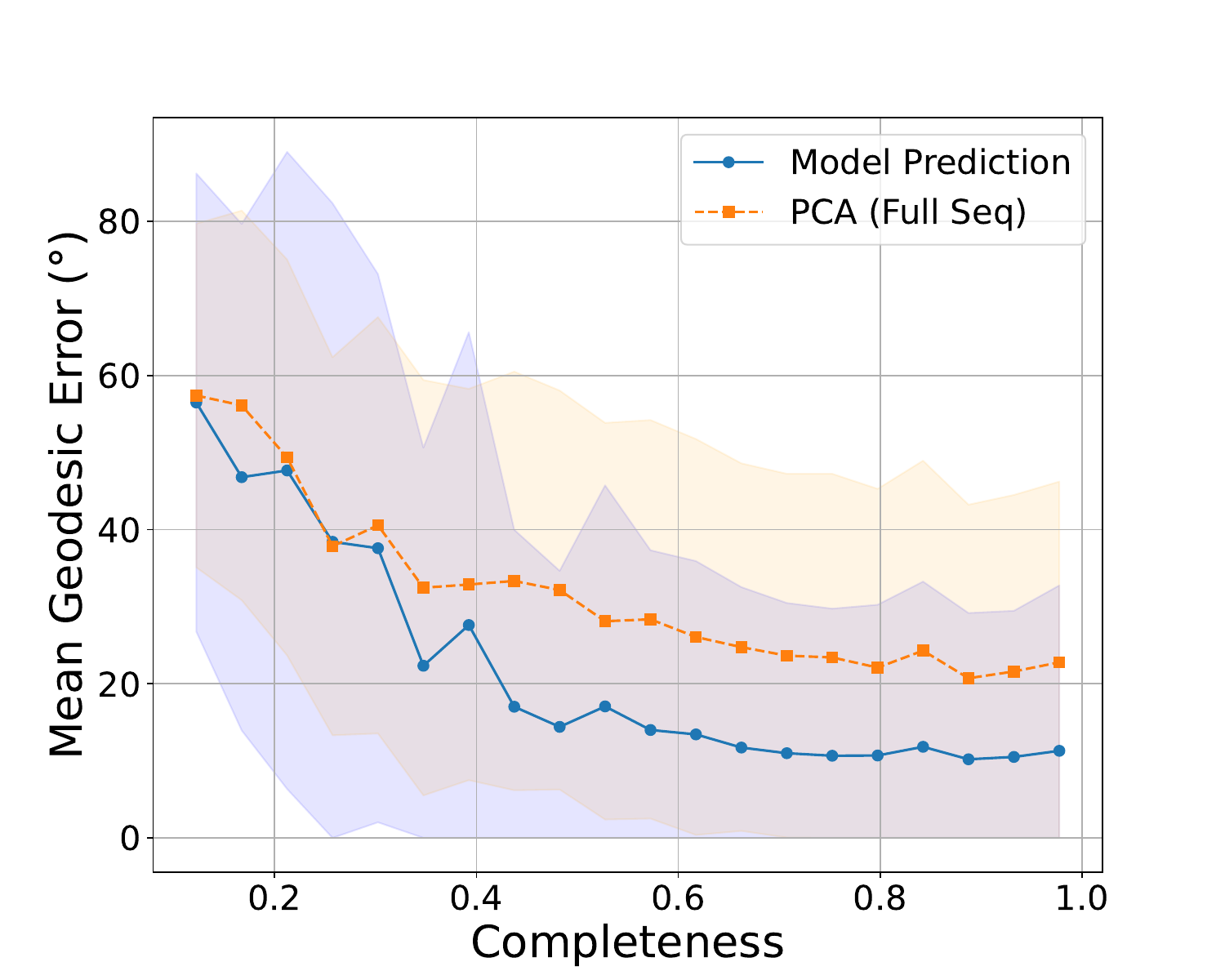}
         \caption{Completeness ($\Phi_r$)}
     \end{subfigure}
     \hfill
     \begin{subfigure}[b]{0.49\linewidth}
         \centering
             \includegraphics[width=0.99\linewidth]{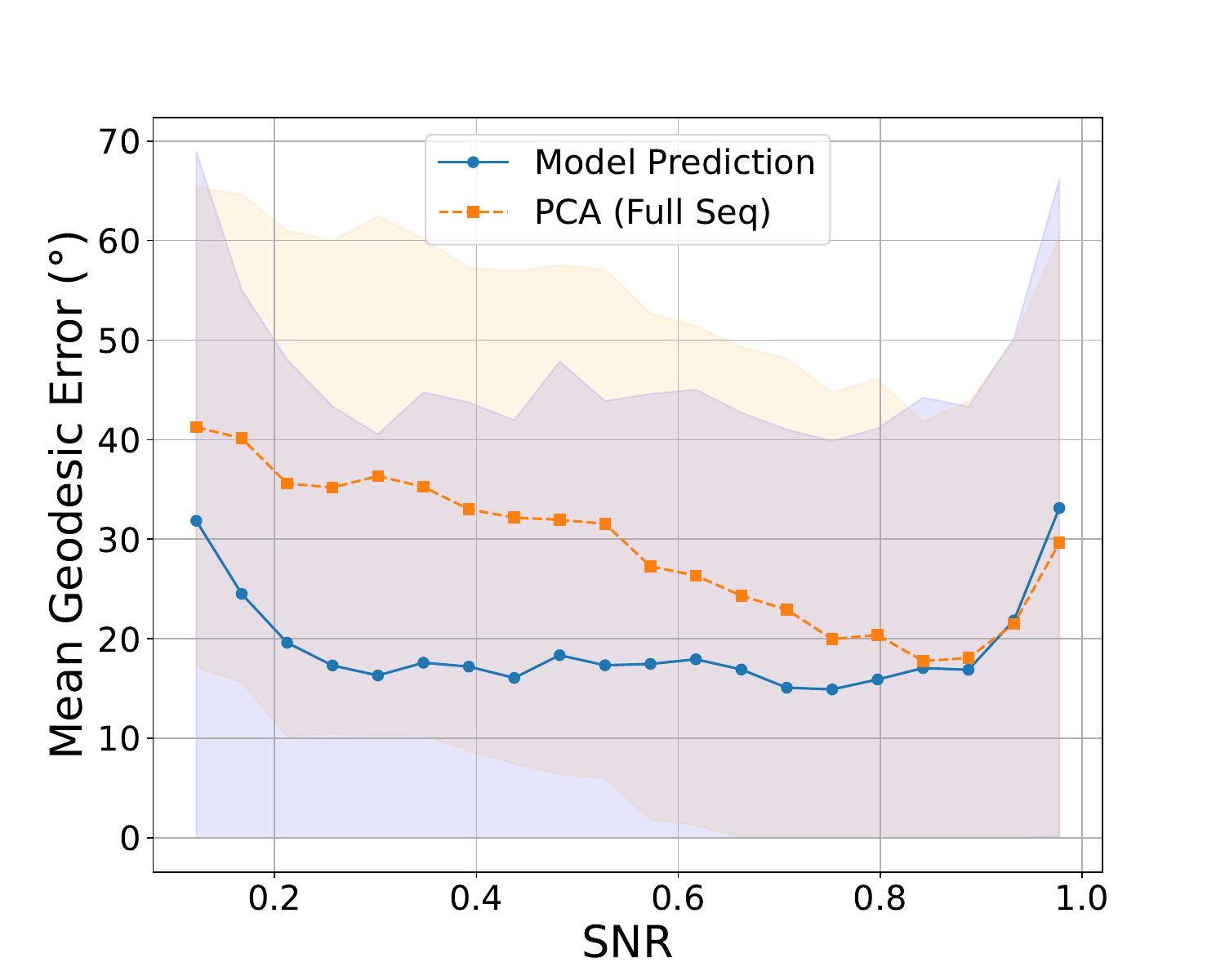}
         \caption{SNR}
     \end{subfigure}
        \caption{
        Comparison between the regression of the normal by the Transformer (Blue) and its estimation by PCA (Orange), for different sequence quality metrics. The plots display average and standard deviation for all test sequences.
        }
        \label{fig: Normal regression comparaison}
        \vspace{-0.15cm}
\end{figure}

\review{A PCA-based estimation of the normal is used as a baseline model for the normal estimation, which is solely done on the position's part of the input sequence.} The eigen vector associated with the minimal eigen value gives the normal to the motion plane. Table~\ref{tab: Transformer vs LSTM vs ResNet vs ViT on the test} shows that the regression models do have a superior performance. \review{Indeed, as the symbol duration might represent a limited part of the sequence, the PCA-based approach is often unreliable.} As seen on Figure~\ref{fig: Normal regression comparaison}, the model normal regression is overall better than the PCA-based one independently of the $\Phi_r$. However, as the SNR goes beyond 0.9 the the PCA-based regression reaches better performance than the model, 0.9 of SNR is quite hard to have when the symbols if further away than 2m. Restricted to the communicative part, the PCA baseline improves but still falls short of the proposed model.

To verify stability, the network was trained using multiple seeds (for the trainset generation); the resulting metrics proved highly consistent, with standard deviations remaining strictly below 0.5\%. For all the following test and ablations study, the train set's seed generation value was chosen as the overall best model on the test set, referred to as the "default model" in the following.

As shown in previous study \cite{10.1145/3495245, enan2022roboticdetectionhumancomprehensiblegestural}, some kinemes are harder to classify. As seen on Table~\ref{tab: f1-score per class default on Test}, our default model can easily detect non-communicative movements and the more complex dynamically the communicative movement is, the easier it is for the model to classify it: the performance on the square and triangle classes is quite lower. If in the input only two segments of the movement are seen, only the corner angle can give the class information, which can be misleading due to the noise, explaining the confusion between the two. On the other hand, the star class is usually the hardest to classify, as it is usually a very small symbol due to its dynamic profile with a lot of acceleration and deceleration phases, which makes it hardly feasible when a big size is sampled. So this class usually presents a very low SNR ratio, preventing a good performance on it.
\begin{table}[ht!]
\vspace{-0.15cm}
\centering
\caption{F1-score over all classes of the Transformer-based architecture}
\begin{tabular}{|l|c|l|c|}
\hline
\textbf{Class} & $F_1^{\text{\tiny classif}} \uparrow$ & \textbf{Class}    & $F_1^{\text{\tiny classif}} \uparrow$ \\ \hline
Square         & 87                                    & Heart             & 95                                    \\ \hline
Circle         & 89                                    & Lemniscate        & 94                                    \\ \hline
Triangle       & 90                                    & Lissajous         & 94                                    \\ \hline
Spiral         & 91                                    & Rose              & 88                                    \\ \hline
Star           & 88                                    & Non-communicative & 95                                    \\ \hline
\end{tabular}
\label{tab: f1-score per class default on Test}
\vspace{-0.15cm}
\end{table}

\begin{figure}[ht!]
     \centering
     \begin{subfigure}[b]{0.49\linewidth}
         \centering
         \includegraphics[width=\linewidth]{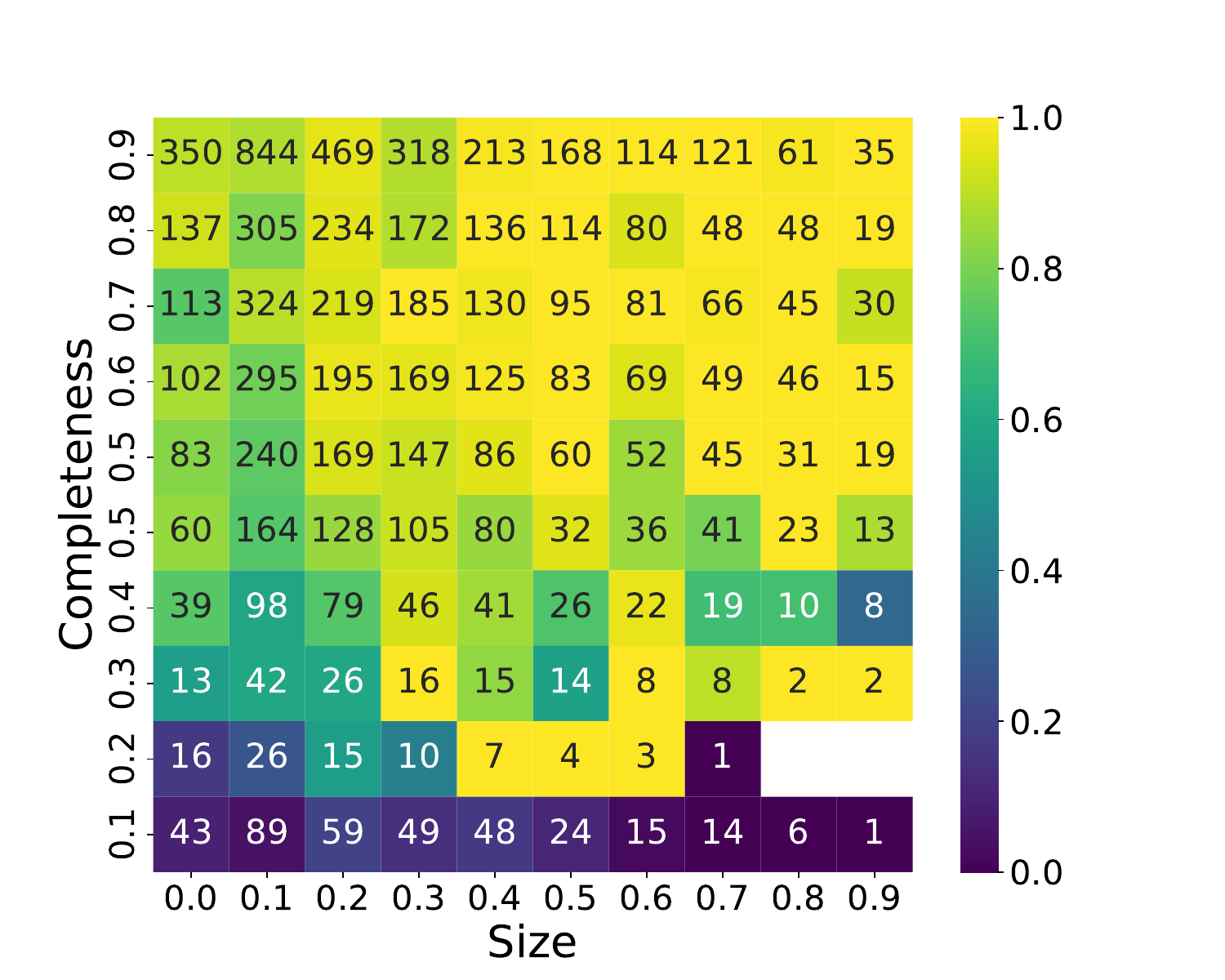}
         \caption{Classification ($F_1^{\text{\tiny classif}} \uparrow  $)}
     \end{subfigure}
     \hfill
     \begin{subfigure}[b]{0.49\linewidth}
         \centering
         \includegraphics[width=\linewidth]{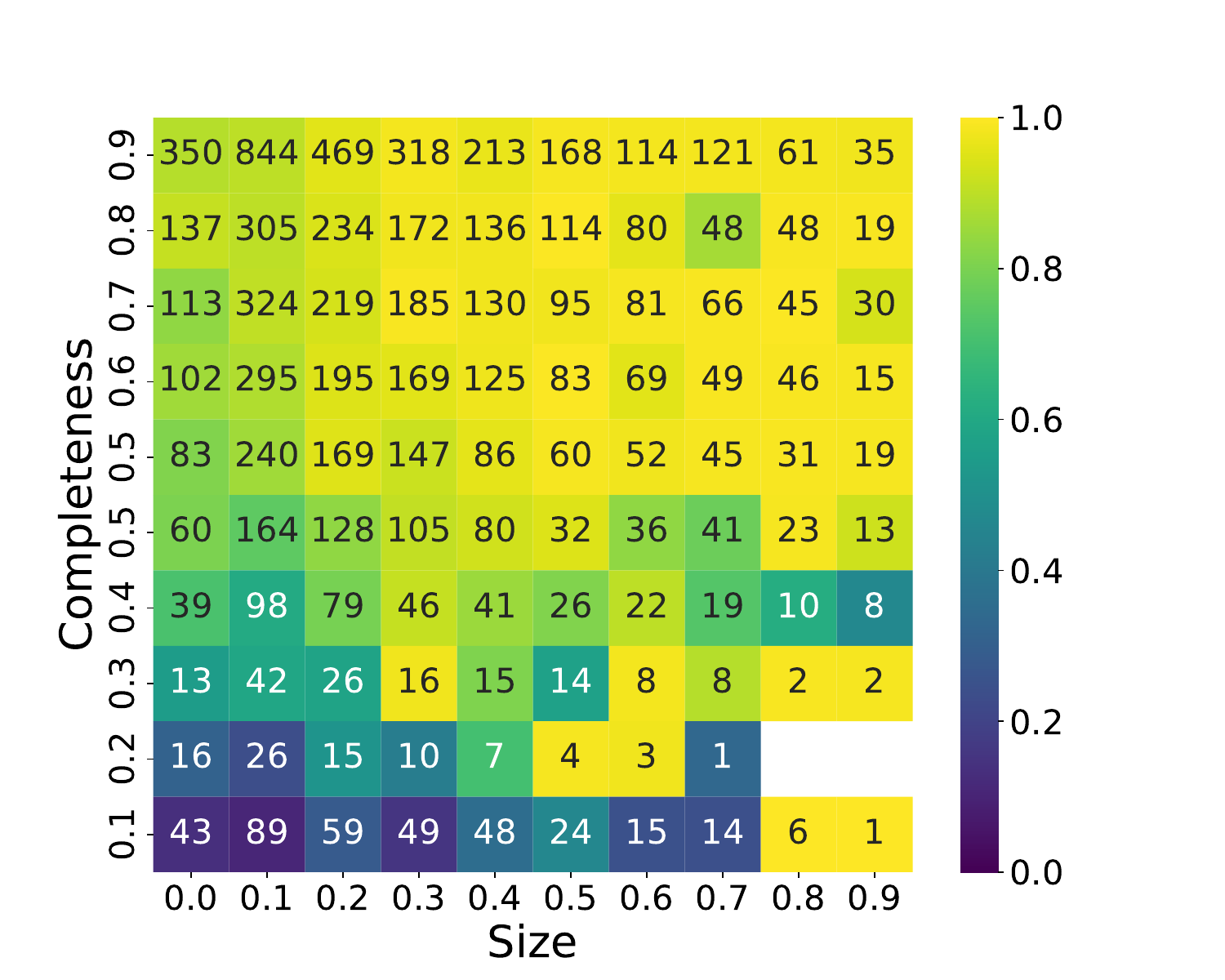}
         \caption{Segmentation ($F_1^{\text{\tiny seg}} \uparrow  $)}
     \end{subfigure}
     \hfill
     \begin{subfigure}[b]{0.49\linewidth}
         \centering
             \includegraphics[width=\linewidth]{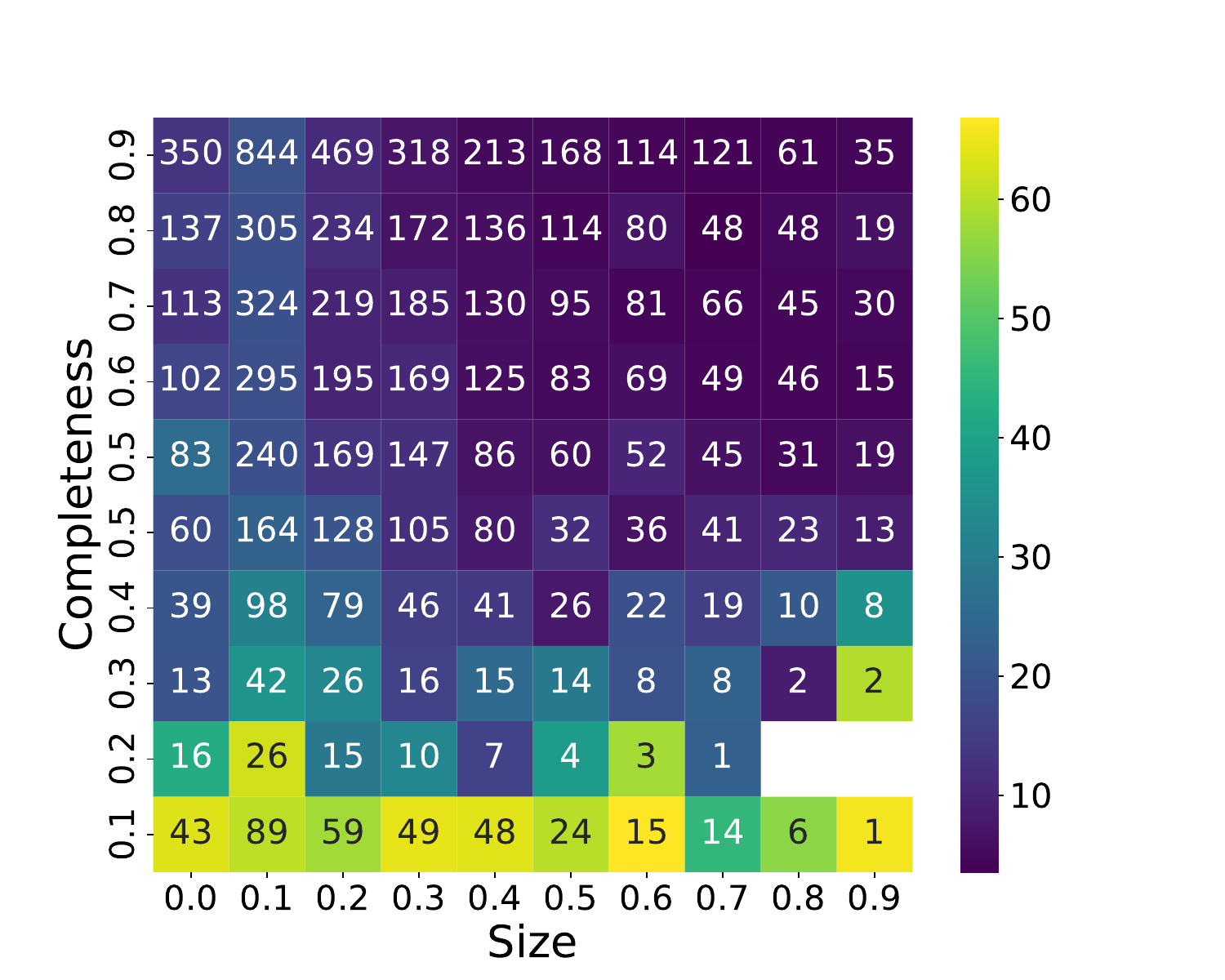}
         \caption{Normal regress. (Mean $\mathcal{E}_G \downarrow$)}
     \end{subfigure}
     \begin{subfigure}[b]{0.49\linewidth}
         \centering
             \includegraphics[width=\linewidth]{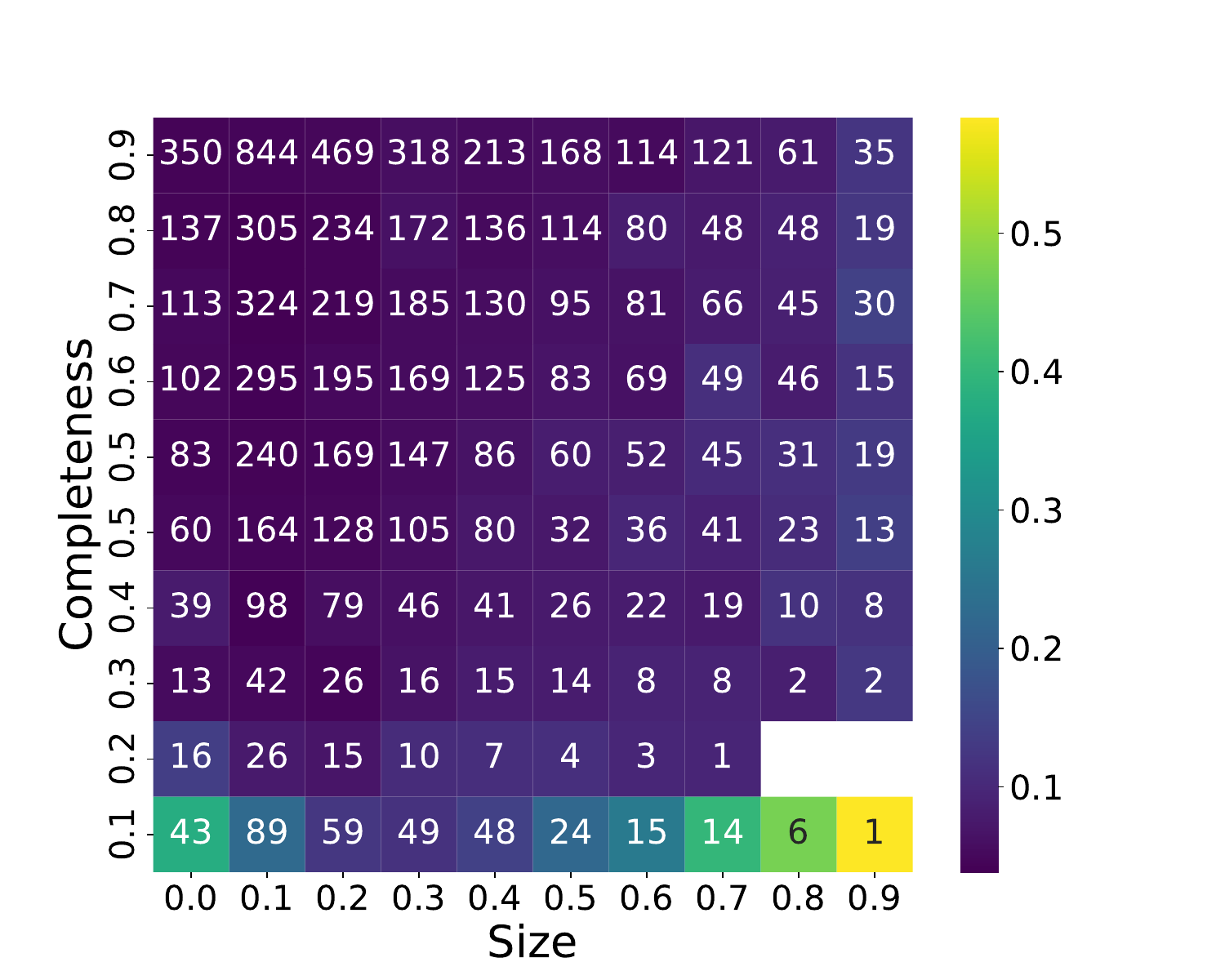}
         \caption{Size regression (MAE $\downarrow$)}
     \end{subfigure}
        \caption{\review{Heatmaps of the default model average performance metrics - on solely the communicative movement - on the test set for the four tasks, depending on Completeness ($\Phi_r$) (vertical) and size (horizontal). The number of samples per combination is shown in the middle of each square.}}
        \label{fig: heatmap default model Completeness / size testset}
         \vspace{-0.15cm}
\end{figure}

\review{So as to give an operating domain of our default model, heatmaps of its performance over the classification, segmentation and regression tasks are shown on Figure~\ref{fig: heatmap default model Completeness / size testset}. Moreover, it is essential to evaluate the performance of the model when the communicative segment is not fully in the input window - which is measured by the Completeness $\Phi_r$ - since in the real swarm, no priors are used to know when the transmitter can send an information through motion, the model has to run at any time. Globally, if we want to maintain a safe communication we should use more than 60$\%$ of $\Phi_r$ i.e., movement lasting more than 7.2s (if the input window is well placed), and should try to have a size over 0.4m. In this operating domain, the model has strong performance with over 85.7$\%$ of $F_1^{\text{\tiny classif}}$, a mean $\mathcal{E}_G$ below 11.5° and a $F_1^{\text{\tiny seg}}$ over 95$\%$. However, the size regression is harder as the movement becomes bigger.}

\begin{table*}[ht!]
\vspace{0.2cm}
\centering
\caption{Ablation on the 3DTrajDecoder's possible inputs. The best result is in bold and the second best is underlined.}
\begin{tabular}{|l|c|c|c|c|c|c|c|}
\hline
\multicolumn{1}{|c|}{On Test} & \begin{tabular}[c]{@{}c@{}}Default\\ (3 inputs)\end{tabular} & Pos    & RelPos & Orient & Pos + RelPos & Pos + Orient & RelPos + Orient \\ \hline
$F_1^{\text{\tiny classif}}$       $\uparrow$               & {\ul 91.13}     & 71.69  & 80.14  &    69.23    & 89.02        & 88.33        & \textbf{93.55}  \\ \hline
$F_1^{\text{\tiny seg}}$    $\uparrow$              & {\ul 94.16}     & 75.16  & 83.12  &    73.54    & 92.12        & 91.15        & \textbf{96.61}  \\ \hline
MAE (m)        $\downarrow$               & \textbf{0.057}  & 0.076  & 0.0625 &   0.1196     & 0.0599       & 0.0767       & {\ul 0.0581}    \\ \hline
MSE (m)         $\downarrow$               & \textbf{0.0061} & 0.0116 & 0.008  &    0.0282    & 0.0068       & 0.0109       & \textbf{0.0061} \\ \hline
$\mathcal{E}_G$ mean(°)  $\downarrow$       & {\ul 13.39}     & 24.32  & 20.02  &   57.34     & 14.02        & 23.63        & \textbf{10.95}  \\ \hline
$\mathcal{E}_G$ median (°) $\downarrow$     & {\ul 5.99}      & 12.77  & 9.87   &   53.95     & 6            & 12.07        & \textbf{4.88}   \\ \hline
Accuracy (\textless{}10°)   $\uparrow$              & {\ul 69.7}      & 39.3   & 50.6   &   3     & 70.6         & 41.9         & \textbf{77.6}   \\ \hline
Accuracy (\textless{}30°)   $\uparrow$              & {\ul 90.9}      & 77.3   & 83.1   &   21.7     & 90           & 78.9         & \textbf{92.9}   \\ \hline
\end{tabular}
\label{tab: comparaison inputs on the test set}
\end{table*}

\begin{figure}[ht!]
     \centering
     \begin{subfigure}[b]{\linewidth}
         \centering
         \includegraphics[width=0.95\linewidth]{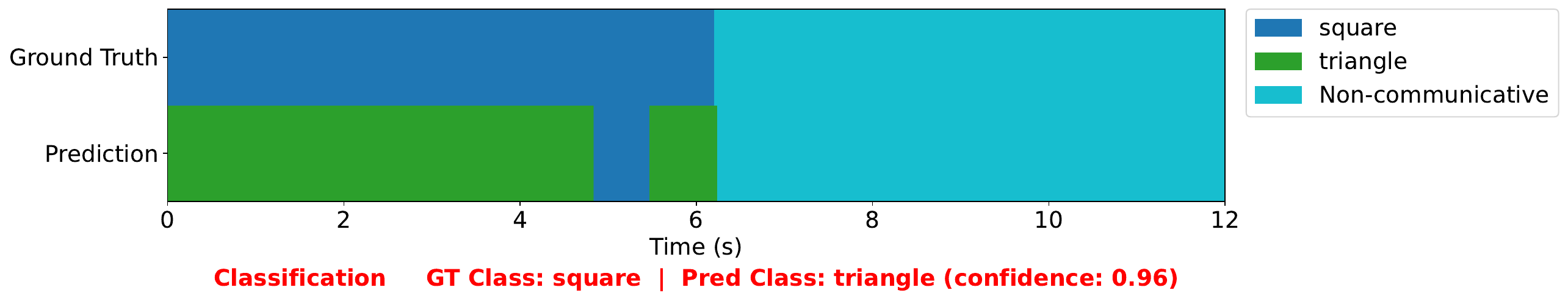}
         \caption{Segmentation of a square (size: 0.308m)}
         \label{fig: segmentation a}
     \end{subfigure}
     \begin{subfigure}[b]{\linewidth}
         \centering
         \includegraphics[width=0.95\linewidth]{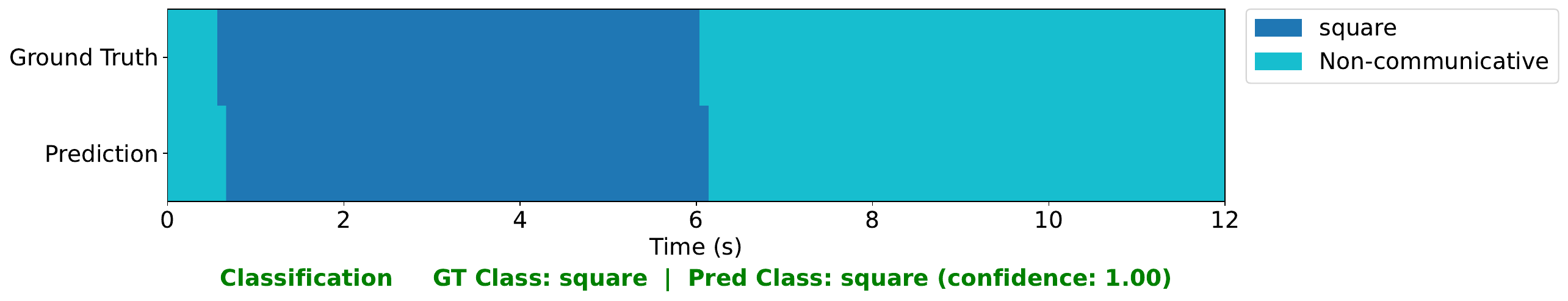}
         \caption{Segmentation of a square (size: 0.096m)}
     \end{subfigure}
     \begin{subfigure}[b]{\linewidth}
         \centering
         \includegraphics[width=0.95\linewidth]{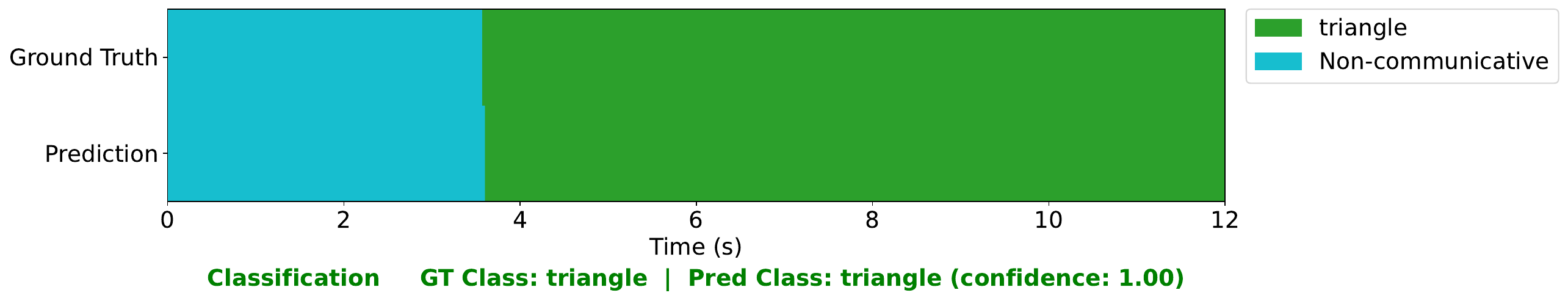}
         \caption{Segmentation of a triangle (size: 0.447m)}
     \end{subfigure}
        \caption{Segmentation and classification results of the model for different input sequences.}
        \label{fig: segmentation}
\end{figure}

\last{Three segmentation outputs corresponding to different input sequences are shown on the figure \ref{fig: segmentation}. The figure \ref{fig: segmentation a} showcases the square and triangle confusion, as solely the first two segments are seen. The two other results show the good performance of the model, even when the movement covers less than half of the input window or that the input sequence start with non-communicative movements.}

\subsection{Ablation on the Transformer heads}
Unlike \cite{enan2022roboticdetectionhumancomprehensiblegestural}, which trained solely a classifier, we trained a multi-task model so as to transmit more information in a single movement. As training a multi-task model usually impacts its performance, we compared our default model - with all the heads trained - to other models with less heads activated. Table~\ref{tab: head activation on test} displays the performance comparison. 
Training solely the classification head does give a small improvement. However, training both the classification and segmentation provides the best performance across both tasks, with an improvement around 1$\%$. \review{The superior size regression achieved by the specifically trained model proves that the default architecture, despite its already strong performance, still has room for further refinement.}
On the normal regression task, training a specific model does not improve the default model. Indeed, the other tasks such as segmentation probably helps create better features, that the normal regression head may naturally use to focus on the communicative segment in the input window.

\begin{table}[ht]
\centering
\caption{Ablation on the Transformer heads. The best result is in bold and the second best is underlined.}
\begin{tabular}{|l|c|c|c|c|c|}
\hline
\multicolumn{1}{|c|}{{\ul \textit{On test}}}                        & \begin{tabular}[c]{@{}c@{}}Default\\ (all \\heads)\end{tabular} & Classif    & \begin{tabular}[c]{@{}c@{}}Classif\\ +\\ Seg\end{tabular} & \begin{tabular}[c]{@{}c@{}}Classif\\ +\\ Size\end{tabular} & \begin{tabular}[c]{@{}c@{}}Classif\\ +\\ Normal\end{tabular} \\ \hline
$F_1^{\text{\tiny classif}}$      $\uparrow$                                                        & 91.13                                                         & {\ul 91.79 }     & \textbf{92.12}                                            & 91.49                                                      & 88.42                                                        \\ \hline
$F_1^{\text{\tiny seg}}$       $\uparrow$                                                 & {\ul 94.16   }                                                     & X          & \textbf{95.57}                                            & X                                                          & X                                                            \\ \hline
MAE (m)      $\downarrow$                                                         & {\ul 0.057    }                                                     & \textbf{X} & X                                                         & \textbf{0.044}                                            & X                                                            \\ \hline
MSE (m)      $\downarrow$                                                         & {\ul 0.006    }                                                   & \textbf{X} & X                                                         & \textbf{0.004}                                            & X                                                            \\ \hline
Mean $\mathcal{E}_G$(°) $\downarrow$    & \textbf{13.39}                                                & X          & X                                                         & X                                                          & {\ul 14.25        }                                                \\ \hline
Median $\mathcal{E}_G$(°) $\downarrow$   & \textbf{5.99}                                                 & X          & X                                                         & X                                                          & {\ul 6.39    }                                                     \\ \hline
Acc (\textless{}10°)   $\uparrow$                                       & \textbf{69.7}                                                 & X          & X                                                         & X                                                          & {\ul 68.4   }                                                      \\ \hline
Acc (\textless{}30°)    $\uparrow$                                  & \textbf{90.9}                                                 & X          & X                                                         & X                                                          & {\ul 89.7  }                                                       \\ \hline
\end{tabular}
\label{tab: head activation on test}
\end{table}

\subsection{Ablation on the possible inputs for the 3DTrajDecoder}

The models can take as input different values given by the pose estimator (see Table~\ref{tab: comparaison inputs on the test set}): the 3D position (Pos), the 3D orientation (Orient) and / or the relative position (RelPos, calculated in reference to the first position of the input sequence). Although redundant, the latter stabilizes drastically the learning of the normal prediction. It also improves the other tasks such as classification and segmentation. Six different combinations of inputs are compared in Table~\ref{tab: comparaison inputs on the test set}. \review{It shows that single-input models provide insufficient information, while dual-input combinations consistently improve performance across tasks. Although omitting absolute position yields higher average test-set metrics, the default three-input model generalizes better, particularly outperforming the RelPos+Orient model on real-world datasets. This generalization gap is most prominent in out-of-distribution size regression: absolute position directly indicates kineme size, whereas relative position is hindered if a sequence begins with non-communicative movements. Ultimately, position data (either absolute or relative) is crucial, as relying solely on orientation results in the poorest performance for both size and normal regression tasks.}

\section{Real-world Experiments}
\review{Following simulation-based validation of the operating domain, we validated the system's practical viability in real-world experiments. As the model was trained entirely on synthetic data, our findings emphasize its effective sim-to-real generalizability.}

\begin{table*}[ht]
\vspace{0.15cm}
\centering
\caption{Performance comparison the default model on the real trajectories using either the perfect minsnap one, or the one actually captured from the MoCAP. The testset is solely composed of the 88 centered communicative segments with the same amount of hover added at the start and end, with the indicated time offset applied. The best result is in bold and the second best is underlined.}
\begin{tabular}{|l|ccccccc|ccccccc|}
\hline
\textit{}                 & \multicolumn{7}{c|}{\begin{tabular}[c]{@{}c@{}}Perfect\\ minsnap\\ trajectories\end{tabular}}                  & \multicolumn{7}{c|}{\begin{tabular}[c]{@{}c@{}}Real\\ Controller\end{tabular}}                               \\ \hline
Time offset used          & -9    & -6           & -3            & \cellcolor[HTML]{C0C0C0}0              & +3             & +6    & +9    & -9    & -6    & -3             & \cellcolor[HTML]{C0C0C0}0             & +3             & +6         & +9    \\ \hline
$F_1^{\text{\tiny classif}}$   $\uparrow$                & 15.04 & 80.83        & 95.84         & \cellcolor[HTML]{C0C0C0}{\ul 97.22}    & \textbf{97.3}  & 87.04 & 32.35 & 11.49 & 73.48 & 92.10          & \cellcolor[HTML]{C0C0C0}{\ul 92.45}   & \textbf{94.26} & 82.22      & 11.92 \\ \hline
$F_1^{\text{\tiny seg}}$     $\uparrow$         & 31.04 & 87.66        & 95.37         & \cellcolor[HTML]{C0C0C0}{\ul 97.03}    & \textbf{97.06} & 96.03 & 36.69 & 22.01 & 76.15 & 86.22          & \cellcolor[HTML]{C0C0C0}{\ul 87.02}   & \textbf{92.91} & 86.14      & 19.95 \\ \hline
MAE (m)    $\downarrow$                & 0.17  & 0.108        & 0.073         & \cellcolor[HTML]{C0C0C0}\textbf{0.059} & {\ul 0.066}    & 0.091 & 0.125 & 0.278 & 0.156 & {\ul 0.075}    & \cellcolor[HTML]{C0C0C0}0.08          & \textbf{0.074} & 0.091      & 0.187 \\ \hline
MSE (m)      $\downarrow$              & 0.052 & 0.02         & 0.008         & \cellcolor[HTML]{C0C0C0}\textbf{0.006} & {\ul 0.007}    & 0.014 & 0.029 & 0.127 & 0.047 & \textbf{0.008} & \cellcolor[HTML]{C0C0C0}0.01          & {\ul 0.009}    & 0.015      & 0.051 \\ \hline
Mean $\mathcal{E}_G$ (°) $\downarrow$  & 33.14 & 5.6          & 4.01          & \cellcolor[HTML]{C0C0C0}\textbf{3.41}  & {\ul 3.57}     & 7.27  & 39.88 & 45.5  & 11.3  & 5.55           & \cellcolor[HTML]{C0C0C0}\textbf{3.49} & {\ul 5.14}     & 9.78       & 44.56 \\ \hline
Median $\mathcal{E}_G$ (°) $\downarrow$  & 20.65 & 4.61         & 3.58          & \cellcolor[HTML]{C0C0C0}\textbf{3.01}  & {\ul 3.25}     & 4.83  & 26.66 & 36.64 & 7.61  & 5.08           & \cellcolor[HTML]{C0C0C0}\textbf{3.08} & {\ul 4.54}     & 7.43       & 34.13 \\ \hline
Accuracy (\textless{}10°) $\uparrow$ & 22.7  & 88.6         & \textbf{98.9} & \cellcolor[HTML]{C0C0C0}\textbf{98.9}  & {\ul 97.7}     & 90.9  & 29.5  & 15.9  & 62.5  & {\ul 89.9}     & \cellcolor[HTML]{C0C0C0}\textbf{98.9} & 89.8           & 71.6       & 15.9  \\ \hline
Accuracy (\textless{}30°) $\uparrow$ & 62.5  & \textbf{100} & \textbf{100}  & \cellcolor[HTML]{C0C0C0}\textbf{100}   & \textbf{100}   & 97.7  & 53.4  & 44.3  & 95.5  & \textbf{100}   & \cellcolor[HTML]{C0C0C0}\textbf{100}  & \textbf{100}   & {\ul 96.6} & 46.6  \\ \hline
\end{tabular}
\label{tab: comp minsnap vs real controller on real data center communicative, only 88 communicative segments min-size-0-4, with timeoffset}
\vspace{-0.10cm}
\end{table*}

\subsection{Real-world setup}
\label{sec: realdata creation}
\begin{figure}[ht]
    \vspace{-0.2cm}
     \centering
     \begin{subfigure}[b]{0.49\linewidth}
         \centering
         \includegraphics[width=0.99\linewidth]{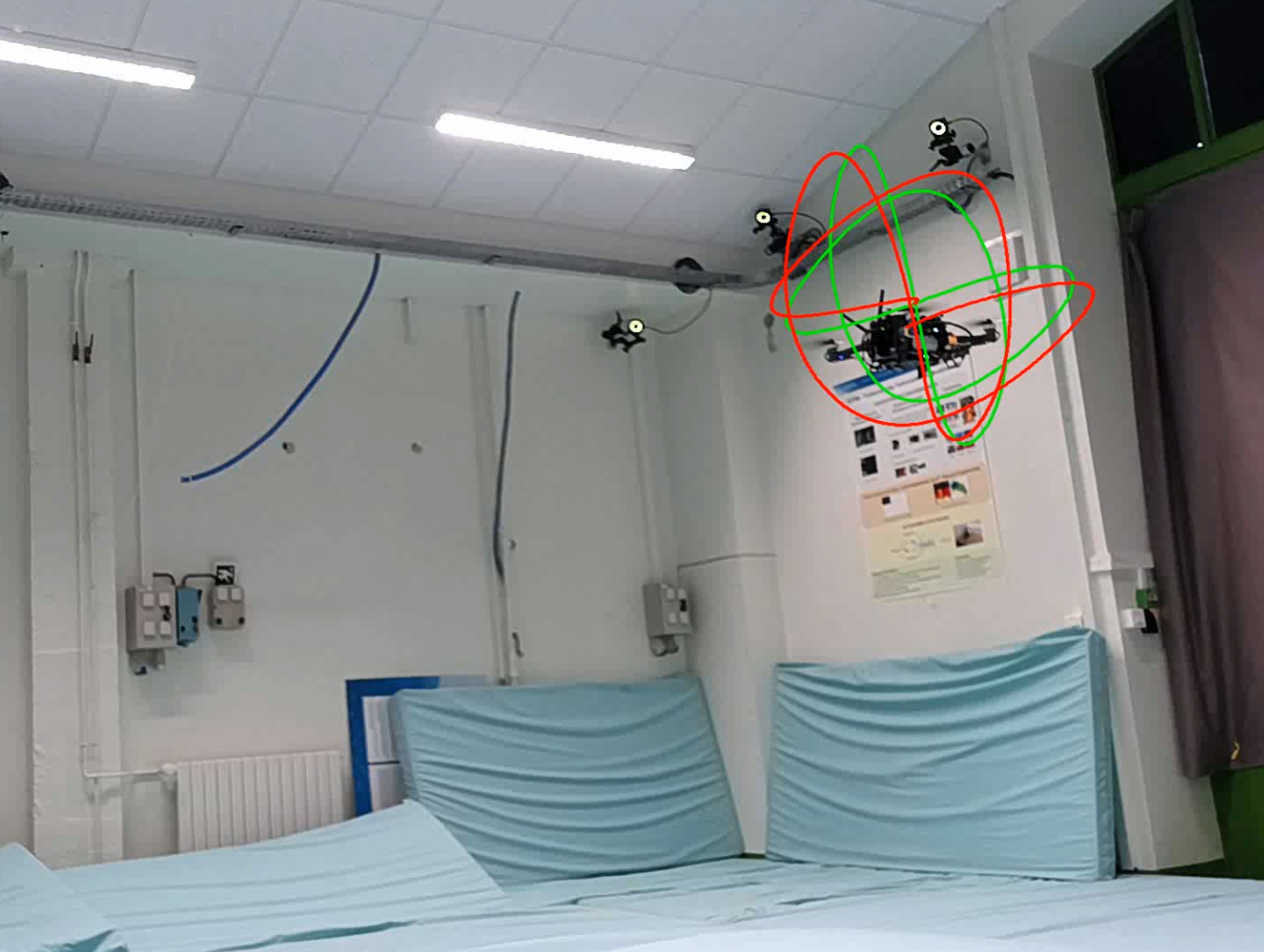}
         \caption{Custom drone communicating (reference trajectory (green), controller-made one (red))}
     \end{subfigure}
     \hfill
     \begin{subfigure}[b]{0.49\linewidth}
         \centering
         \includegraphics[width=0.8\linewidth]{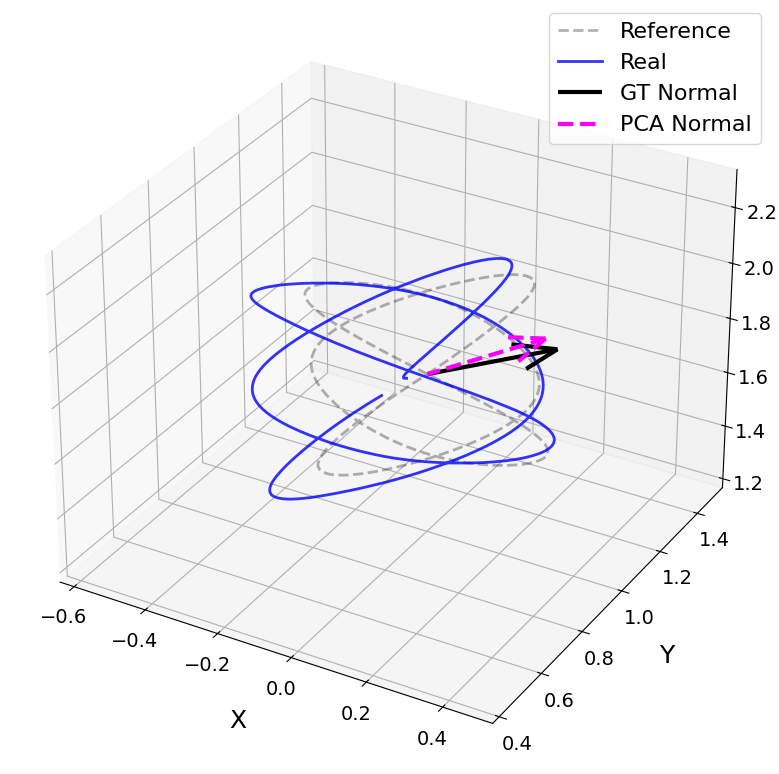}
         \caption{Comparison of the reference trajectory and controller-made one for a lissajous}
     \end{subfigure}
     \caption{Real image from the drone following a lissajous (left) and comparison of the reference and 
     controller trajectories (right). The planar normal GT and PCA-estimated are shown.}
     \label{fig: real trajectory example}
      \vspace{-0.3cm}
 \end{figure}

While our synthetic dataset consists of isolated 12-second trajectory windows, evaluating the model on physical hardware necessitates long-horizon, continuous flight sequences. Rather than executing a single gesture, the real-world flight comprises a suite of distinct communicative phases interleaved with non-communicative transit movements and hover phases lasting 6s at the start and the end of each communicative segment. Consistent with the synthetic generation and following our model's operating domain for a safe communication, the shape class, physical size, spatial orientation (normal vector), and total execution time for each communicative phase are randomly sampled, taking into account the operating domain, see Figure \ref{fig: heatmap default model Completeness / size testset}. Deploying on physical hardware introduces strict safety requirements. \review{Trajectories are heavily constrained within a predefined safe spatial bounding box (4x4x2.5m) to prevent collisions with our flying zone area boundaries. }
\review{Minimum snap sequences that violate geometric or dynamic constraints are discarded and regenerated. For real flights, valid optimized polynomials are discretized at 20 Hz to provide reference setpoints for the MoCap-based tracking controller.} Following the data generation pipeline, shown in Figure~\ref{fig: dataset pipeline generation}, 88 communicative segments are generated. An example of a controller reference trajectory and the one actually completed is illustrated in Figure~\ref{fig: real trajectory example}.

A rosbag containing the MoCAP pose, \review{and the image seen from a wheeled-robot, to be exploited in future tests are saved.
From this rosbag a file containing the 30Hz downsampled MoCAP information and the labels is generated, 
synthetic observation noise is added following the synthetic dataset generation pipeline}. Over the almost 30min of recorded flight, the 88 kinemes are saved so as to be centered in the 12s input window, as they can last from 7.2 to 12s. Moreover, more than 50 samples of non-communicative movements - done between each kinemes - are also saved. \review{Due to the nature of the controller, the supposed normal of the communicative phase and its size are not fully respected. When evaluating the executed trajectories against the reference setpoints, we observe a mean geodesic error of 3.7° and a mean size error of 0.072m (where kineme size is sampled from $\mathcal{U}(0.4,1)m$).}%

\subsection{Real-world results}

During training, the model saw a minsnap trajectory with some simulated controller noise on the waypoints. 
In the real case, we need to evaluate how the trajectory generated by the real controller impacts the performance over the 88 communicative movements. Table~\ref{tab: comp minsnap vs real controller on real data center communicative, only 88 communicative segment min-size-0-4} shows the performance of the default model on the perfect communicative movements and the ones actually made. Overall, classification, segmentation and size regression are impacted by the controller-due error, but the performance remains quite good, as we restrict the real trajectories within the operating domain of our model, showcasing a real use-case. \review{Moreover, the syntheticaly trained model does also have a strong performance on the 50 non-communicative movements with more than 92\% $F_1^{\text{\tiny classif}}$.}

\begin{table}[ht]
\vspace{0.15cm}
\centering
\caption{Performance comparison of the default model on the real trajectories, using either the perfect minsnap one, or the one actually captured from the MoCAP. The testset is solely composed of the 88 centered communicative segments}
\begin{tabular}{|l|c|c|}
\hline
\textit{\begin{tabular}[c]{@{}l@{}}Real data\\ Min-size-0.4\end{tabular}} & \begin{tabular}[c]{@{}c@{}}Perfect\\ minsnap\\ trajectories\end{tabular} & \begin{tabular}[c]{@{}c@{}}Real\\ Controller\end{tabular} \\ \hline
$F_1^{\text{\tiny classif}}$      $\uparrow$                                                            & \textbf{97.22}                                                           & 92.45                                                     \\ \hline
$F_1^{\text{\tiny seg}}$        $\uparrow$                                                      & \textbf{97.03}                                                           & 87.02                                                     \\ \hline
MAE (m)         $\downarrow$                                                          & \textbf{0.0589}                                                          & 0.0802                                                    \\ \hline
MSE (m)          $\downarrow$                                                          & \textbf{0.0061}                                                          & 0.0099                                                    \\ \hline
Mean $\mathcal{E}_G$   (°)    $\downarrow$                                                 & \textbf{3.41}                                                            & 4.72                                                      \\ \hline
Median $\mathcal{E}_G$    (°)   $\downarrow$                                               & \textbf{3.01}                                                            & 4.12                                                      \\ \hline
Accuracy (\textless{}10°)      $\uparrow$                                           & \textbf{98.9}                                                            & 
95.5
\\ \hline
Accuracy (\textless{}30°)      $\uparrow$                                           & \textbf{100}                                                             & \textbf{100}                                              \\ \hline
\end{tabular}
\vspace{-0.15cm}
\label{tab: comp minsnap vs real controller on real data center communicative, only 88 communicative segment min-size-0-4}
\end{table}

\review{As real data is to be used sequentially, and no priors are assumed to know when the transmitter can send an information through motion, the model will not be used on every timestamps. So we want to test its performance over some time offsets w.r.t. the perfectly centered communicative segment, so as to simulate a sliding window over the sequentially used data.} On Table~\ref{tab: comp minsnap vs real controller on real data center communicative, only 88 communicative segments min-size-0-4, with timeoffset}, we show the performance on the 'perfect' real trajectories and the ones seen through the MoCAP, with different time offsets between $\pm9s$. 
\review{As the input window aggregates the information on 12s, all those time offsets show a wide range of defects, which might appear when a sliding window is used.} Overall the model is quite robust to $\pm 3s$ offsets, maintaining a pretty good performance over all the tasks, even sometimes exceeding the performance of the perfectly aligned one. However, when big time offset are used, less than 50$\%$ of Completeness ($\Phi_r$) - i.e., more than $\pm 6s$, there is quite a big drop in performance, which is also observed on synthetic data.

\section{Conclusion}

This paper presents a novel approach to UAV swarm communication, addressing the critical vulnerabilities of traditional radio frequency systems in denied environments. By conceptualizing 3D flight trajectories as a structured kinematic alphabet, we demonstrated the feasibility of transmitting complex semantic information - such as command class, magnitude and spatial directionality - , strictly through a modulated movement. Using a sensor-agnostic pose estimator - modeled here with RGB-level observation noise - to capture the transmitter's trajectory, we decoded this visual language using 3DTrajDecoder. This multi-task transformer simultaneously classifies, segments, and regresses the size and normal of kinemes directly from pose sequences. The architecture is validated across both synthetic and real-world datasets. 

\review{Future works will focus on the real-time deployment of the complete end-to-end pipeline, from RGB input directly to message decoding. This includes integrating the 3DTrajDecoder within a ROS framework for live inference on a custom quadrotor platform, using an RGB pose estimator to process the RGB stream. Ultimately, we aim to evaluate the communication's robustness in fully decentralized, multi-agent exploration scenarios particularly under challenging conditions such as dynamic backgrounds, variable lighting, and continuous receiver motion. Our ablation study shows that full pose estimation is not a strict requirement for the 3DTrajDecoder. A highly efficient alternative relies simply on positional tracking via standard object detection, leveraging a priori knowledge of the transmitter's scale. Moreover, since transmitting drones typically localize receivers using their own onboard estimators, they can dynamically adjust their communicative movements to maintain reliable transmission within the model's operating domain.}

\bibliographystyle{IEEEtran}
\bibliography{IEEEabrv,ifacconf.bib}

\begin{thebibliography}{10}
\providecommand{\url}[1]{#1}
\csname url@rmstyle\endcsname
\providecommand{\newblock}{\relax}
\providecommand{\bibinfo}[2]{#2}
\providecommand\BIBentrySTDinterwordspacing{\spaceskip=0pt\relax}
\providecommand\BIBentryALTinterwordstretchfactor{4}
\providecommand\BIBentryALTinterwordspacing{\spaceskip=\fontdimen2\font plus
\BIBentryALTinterwordstretchfactor\fontdimen3\font minus
  \fontdimen4\font\relax}
\providecommand\BIBforeignlanguage[2]{{%
\expandafter\ifx\csname l@#1\endcsname\relax
\typeout{** WARNING: IEEEtran.bst: No hyphenation pattern has been}%
\typeout{** loaded for the language `#1'. Using the pattern for}%
\typeout{** the default language instead.}%
\else
\language=\csname l@#1\endcsname
\fi
#2}}

\bibitem{SaskaGestureHSI2025}
V.~Krátký, G.~Silano, M.~Vrba, C.~Papaioannidis, I.~Mademlis, R.~Pěnička,
  I.~Pitas, and M.~Saska, ``Gesture-controlled aerial robot formation for
  human-swarm interaction in safety monitoring applications,'' \emph{IEEE
  Robotics and Automation Letters}, vol.~10, no.~8, p. 8244–8251, Aug. 2025.

\bibitem{yu2025electronic}
A.~Yu, I.~Kolotylo, H.~A. Hashim, and A.~E. Eltoukhy, ``Electronic warfare
  cyberattacks, countermeasures and modern defensive strategies of uav
  avionics: a survey,'' \emph{IEEE Access}, 2025.

\bibitem{9836151}
J.~Horyna, V.~Walter, and M.~Saska, ``Uvdar-com: Uv-based relative localization
  of uavs with integrated optical communication,'' in \emph{2022 International
  Conference on Unmanned Aircraft Systems (ICUAS)}, 2022, pp. 1302--1308.

\bibitem{baltatzis2024neuralsignactorsdiffusion}
V.~Baltatzis, R.~A. Potamias, E.~Ververas, G.~Sun, J.~Deng, and S.~Zafeiriou,
  ``Neural sign actors: A diffusion model for 3d sign language production from
  text,'' 2024.

\bibitem{10.1145/3495245}
M.~Fulton, C.~Edge, and J.~Sattar, ``Robot communication via motion: A study on
  modalities for robot-to-human communication in the field,'' \emph{J.
  Hum.-Robot Interact.}, vol.~11, no.~2, Feb. 2022.

\bibitem{10143200}
J.~Chen, J.~Wang, Q.~Yuan, and Z.~Yang, ``Cnn-lstm model for recognizing
  video-recorded actions performed in a traditional chinese exercise,''
  \emph{IEEE Journal of Translational Engineering in Health and Medicine},
  vol.~11, pp. 351--359, 2023.

\bibitem{wu2025torchspatiallocationencodingframework}
N.~Wu, Q.~Cao, Z.~Wang, Z.~Liu, Y.~Qi, J.~Zhang, J.~Ni, X.~Yao, H.~Ma, L.~Mu,
  S.~Ermon, T.~Ganu, A.~Nambi, N.~Lao, and G.~Mai, ``Torchspatial: A location
  encoding framework and benchmark for spatial representation learning,'' 2025.

\bibitem{Animal2Swarm2023}
H.~Duan, M.~Huo, and Y.~Fan, ``From animal collective behaviors to swarm
  robotic cooperation,'' \emph{National Science Review}, vol.~10, no.~5, 2023.

\bibitem{5354808}
D.~Raghunathan and J.~Baillieul, ``Motion based communication channels between
  mobile robots - a novel paradigm for low bandwidth information exchange,'' in
  \emph{2009 IEEE/RSJ International Conference on Intelligent Robots and
  Systems}, 2009, pp. 702--708.

\bibitem{baillieul2011controltheorymotionbasedcommunication}
J.~Baillieul and K.~Özcimder, ``The control theory of motion-based
  communication: Problems in teaching robots to dance,'' 2011.

\bibitem{nishimura2018active}
H.~Nishimura and M.~Schwager, ``Active motion-based communication for robots
  with monocular vision,'' in \emph{2018 IEEE International Conference on
  Robotics and Automation (ICRA)}.\hskip 1em plus 0.5em minus 0.4em\relax IEEE,
  2018, pp. 2948--2955.

\bibitem{enan2022roboticdetectionhumancomprehensiblegestural}
S.~S. Enan, M.~Fulton, and J.~Sattar, ``Robotic detection of a
  human-comprehensible gestural language for underwater multi-human-robot
  collaboration,'' 2022.

\bibitem{vanc2023communicating}
P.~Vanc, J.~K. Behrens, K.~Stepanova, and V.~Hlavac, ``Communicating human
  intent to a robotic companion by multi-type gesture sentences,'' in
  \emph{2023 IEEE/RSJ International Conference on Intelligent Robots and
  Systems (IROS)}.\hskip 1em plus 0.5em minus 0.4em\relax IEEE, 2023, pp.
  9839--9845.

\bibitem{albergo2025motion}
N.~Albergo, J.~Hwang, D.~Lee, and K.~Han, ``Motion-based communication as a
  language: Formal grammar representation and model-free decoding with dense
  optical flow,'' \emph{ASCE OPEN: Multidisciplinary Journal of Civil
  Engineering}, vol.~3, no.~1, p. 04025010, 2025.

\bibitem{zhou2019continuity}
Y.~Zhou, C.~Barnes, J.~Lu, J.~Yang, and H.~Li, ``On the continuity of rotation
  representations in neural networks,'' in \emph{Proceedings of the IEEE/CVF
  conference on computer vision and pattern recognition}, 2019, pp. 5745--5753.

\bibitem{vaswani2017attention}
A.~Vaswani, ``Attention is all you need,'' \emph{Advances in Neural Information
  Processing Systems}, 2017.

\bibitem{Mueller2015}
M.~W. Mueller, M.~Hehn, and R.~D'Andrea, ``A computationally efficient motion
  primitive for quadrocopter trajectory generation,'' \emph{IEEE Transactions
  on Robotics}, vol.~31, no.~6, pp. 1294--1310, 2015.

\bibitem{he2016deep}
K.~He, X.~Zhang, S.~Ren, and J.~Sun, ``Deep residual learning for image
  recognition,'' in \emph{Proceedings of the IEEE conference on computer vision
  and pattern recognition}, 2016, pp. 770--778.

\bibitem{REY2025103339}
T.~Rey, J.~Moras, A.~Eudes, and A.~Manzanera, ``Real-time visual pose
  estimation: from bop objects to custom drone — a journey,''
  \emph{Mechatronics}, vol. 109, p. 103339, 2025.

\bibitem{stapf2023pvit}
S.~Stapf, T.~Bauernfeind, and M.~Riboldi, ``Pvit-6d: Overclocking vision
  transformers for 6d pose estimation with confidence-level prediction and pose
  tokens,'' \emph{arXiv preprint arXiv:2311.17504}, 2023.

\end{thebibliography}

\end{document}